\pdfoutput=1
\RequirePackage{fix-cm}
\documentclass[twocolumn]{svjour3}          
\smartqed  
\usepackage{graphicx}
\usepackage{multirow}

\begin{document}

\title{A Co-Prime Blur Scheme for Data Security in Video Surveillance
}


\author{Christopher Thorpe
	\and Feng Li
	\and Zijia Li
        \and Zhan Yu
	\and David Saunders
	\and Jingyi Yu 
}


\institute{C. Thorpe, F. Li, Z. Yu, D. Saunders, J. Yu \at
              University of Delaware, Newark, DE 19716 \\
              Tel.: +1-302-831-4445\\
              Fax: +1-302-831-8458\\
              \email{\{thorpe, feli, zyu, saunders, yu\}@cis.udel.edu}           
           \and
           Z. Li \at
           Key Laboratory of Mathematics Mechanization\\
           AMSS, Beijing, China 100190 \\
           \email{lizijia@amss.ac.cn}
}

\date{Received: date / Accepted: date}

\maketitle

\begin{abstract}
This paper presents a novel Coprime Blurred Pair (CBP) model for visual data-hiding for security in camera surveillance. While most previous approaches have focused on completely encrypting the video stream, we introduce a spatial encryption scheme by blurring the image/video contents to create a CBP. Our goal is to obscure detail in public video streams by blurring while allowing behavior to be recognized and to quickly deblur the stream so that details are available if behavior is recognized as suspicious. We create a CBP by blurring the same latent image with two unknown kernels. The two kernels are coprime when mapped to bivariate polynomials in the $z$ domain. To deblur the CBP we first use the coprime constraint to approximate the kernels and sample the bivariate CBP polynomials in one dimension on the unit circle. At each sample point, we factor the 1D polynomial pair and compose the results into a 2D kernel matrix. Finally, we compute the inverse Fast Fourier Transform (FFT) of the kernel matrices to 
recover the coprime kernels and then the latent video stream. It is therefore only possible to deblur the video stream if a user has access to both streams. To improve the practicability of our algorithm, we implement our algorithm using a graphics processing unit (GPU) to decrypt the blurred video streams in real-time, and extensive experimental results demonstrate that our new scheme can effectively protect sensitive identity information in surveillance videos and faithfully reconstruct the unblurred video stream when two blurred sequences are available.

\end{abstract}

\section{Introduction}
\label{intro}

Video surveillance in public spaces has increased dramatically in recent history as a means to deter both terrorism and crime in urban environments. However, concern about the potential for abuse and the general loss of privacy has also grown along with the number of surveillance cameras. In recent years the problem of providing visual data in sensitive environments without impinging on the privacy of the public has been a major research topic in machine vision society. The challenge in protecting surveillance data lies in how to reduce both the number of personally identifiable features and the people with access to these features \cite{yangjie07,yangjie06,yangjie08}.

State-of-the-art information hiding techniques either rely on cryptography (codes and ciphers) to change the structure of data or use steganography to render the message invisible. For visual data, Naor and Shamier \cite{naor1995visual} first introduced the notion of visual cryptography. They also developed a Visual Secret Sharing (VSS) scheme by dividing the secret image among $n$ participants where the image can only be recovered when a sufficient number of participants stack their respective pieces together. VSS can effectively protect the visual data. However, since the data received by each individual participant has been fully encrypted, it provides very little usable information. In the case of video surveillance, this indicates that the data received by low-clearance users would be completely useless.

In this paper we present a novel visual data-hiding technology for data security in video surveillance. Different from previous \textquotedblleft complete\textquotedblright encryption schemes, we introduce a \textquotedblleft partial\textquotedblright data encryption scheme by blurring the image/video contents. Our goal is two fold, to provide anonymity to the public and to limit user access to image streams that have less degrees of blurring or no blurring at all. Our approach makes use of two blurred image streams each of which suppress the appropriate private attributes while retaining the ability to reconstruct an unblurred image if one has access to both image streams simultaneously. Experimental results show that our new scheme can effectively protect sensitive identity information in surveillance videos and reconstruct the original unblurred video stream for people with high security clearance.



\section{Related Work}
\label{rel_work}

Most existing visual data protection schemes fall into the framework of visual cryptography (VC) first proposed by Naor and Shamir \cite{naor1995visual}. The main idea is to partition the data into n pieces called the shares. Only when sufficient number of shares are stacked together, will human eyes recognize the image content. To achieve this goal, the authors superimpose random patterns of dots onto each share to produce noisy-looking images. However, the appearance of these noisy images in and of themselves might lead one to suspect data encryption. To alleviate this problem, newer schemes such as halftoning \cite{zhou2006halftone,myodo2007halftone,wang2006halftone} attempt to separately deal with grayscale and color channels to minimize the loss in contrast . They also produce visually pleasing results for the shares. For example, the work by Zhou et al. \cite{zhou2006halftone} observes that the grayscale component carries the most significant visual cues and they chose to generate halftone shares on a 
second image.

Existing VC solutions, however, are not directly applicable to privacy-protection in visual surveillance. For example, classical VC schemes completely hide the image contents from low-clearance users and the encrypted visual data is completely useless to these users. Ideally, low-clearance users should still be able to access some visual information (e.g., the behavior of the target) although their ability for viewing privacy-sensitive details such as facial features would be constrained. These observations have led to the notion of \textquotedblleft blind vision\textquotedblright for addressing such privacy issues. Recent efforts include privacy-protected face detection \cite{avidan06}, face recognition \cite{erkin2009privacy}, image filtering \cite{hu2007secure}, and image retrieval \cite{shashank2008private}. Along the same line of the surveillance work by Yang et al. \cite{yangjie07,yangjie06,yangjie08} that aim to separately treat privacy vs. non-privacy features, we present a \textquotedblleft partial\
textquotedblright visual encryption scheme to allow low-clearance users to partially analyze the visual data without revealing crucial identity information. A comprehensive study of VC can be found in \cite{weir2010comprehensive}.

A natural \textquotedblleft partial\textquotedblright encryption solution is to strategically blur the imagery data. A blurred image $B$ can be viewed as the convolution of a latent image $L$ with a blur kernel $K$. Tremendous efforts have been focused on solving the \emph{blind} image deconvolution problem in which neither $L$ nor $K$ is known. Since blind deconvolution is an under-constrained problem, state-of-the-art solutions rely on regularization to avoid trivial solutions \cite{shan2008high,xu2010two,Tai11}. Latest approaches attempt to use special priors such as image statistics \cite{fergus06}, edge and gradient distributions \cite{levin2007blind}, kernel sparsity and continuity \cite{jia08}, or color information \cite{joshi2009image} for both kernel estimation and image deconvolution. Despite these advances, robust deconvolution remains as an open problem in image processing and computer vision \cite{levin2009understanding}. However, because accurate deblurring is inherently difficult it is ideal 
for partial visual encryption. The blurry video stream protects identity details while withstanding brute-force deconvolution methods intended to \textquotedblleft decrypt\textquotedblright the data.


Our solution is inspired by recently proposed dual-image deblurring techniques \cite{rav05,yuan2007image,jia08}. These methods use a pair of images captured towards the same scene under different aperture/shutter settings. For example, a blurry/noisy image pair can be captured with different shutter speeds. The image pair helps to estimate the kernel and to reduce the ringing artifacts \cite{yuan2007image} in reconstruction. A dual-blur pair \cite{jia08} captures the scene under different motion blurs. It then estimates both blur kernels by constructing an equally blurred image pair. These methods suggest that the correlation between the images imposes important constraints that are useful for kernel and latent image estimations. In particular, it indicates that a single image/video stream will be impossible to decrypt (e.g., to a low-clearance user) but if both streams are available, they will be easy to decrypt (e.g., to a high-clearance user).

\section{Algorithm Overview}
\label{sys_over}

\begin{figure*}
 \begin{center}
\includegraphics[width=0.7\linewidth]{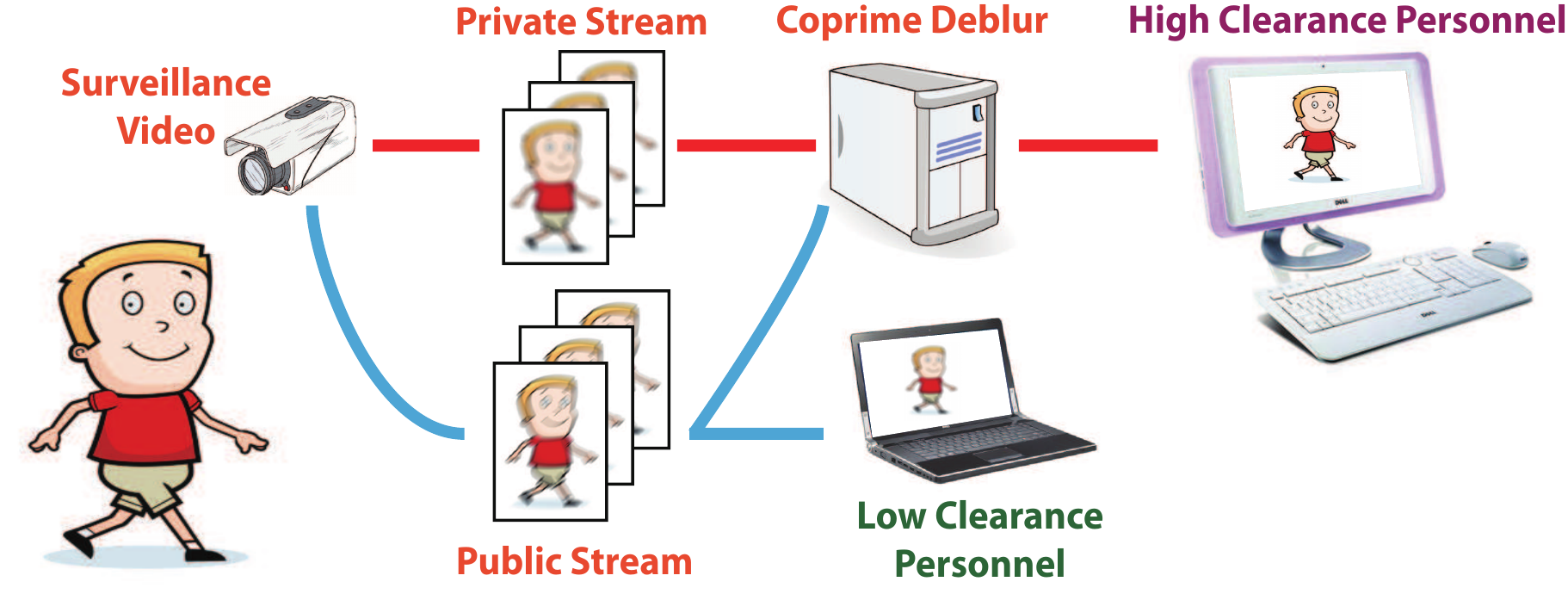}
\end{center}
 \caption{The processing pipeline of our data-hiding technology for data security in video surveillance.}
 \label{fig:systempipeline}
\end{figure*}

Fig. \ref{fig:systempipeline} shows the processing pipeline of our data-hiding technology for data security in video surveillance. We apply coprime blur kernels to the input surveillance video sequence to generate two blurry video streams where all the sensitive information is hidden. One is the public stream, and the other is the private stream. The private video sequence is streamed to high clearance personnel through secure transmission lines. Only high clearance personnel has access to both the private and public streams. Here kernel coprimality means when viewing the blur kernels as 2D polynomials, these 2D polynomials are coprime and the degree of their GCD is 1. By using the coprime blur kernels for sensitive data hiding, we design an efficient and robust algorithm to faithfully reconstruct the ground truth input sequence for the high clearance personnel. We also show that by implementing it on GPUs, we can further improve the computational efficiency of the algorithm to be real-time on a typical 
business workstation. This would make our data-hiding technology practical for everyday surveillance environments.

\textbf{Blur Kernel Selection:} The manner in which the blur kernel is implemented in our system is important because it impacts the security of the system and the privacy due to the blurring. The key is to use very large blur kernels when compared to the size of the input image sequence. The larger the kernel size, the more unknown variables there are to solve in an inverse problem, and thus the harder it is to decrypt the surveillance video. We therefore randomly construct the blur kernels of large size, and use different kernels for each new frame in the stream. The end users need not to know when the blur kernel has changed because they have no means of deblurring the image with only one image. Users with access to both image streams also need not to know when a blur kernel changes or even the blur kernel itself because the deblurring processes simply relies on possessing two images simultaneously. However, changing the blur kernel frequently may affect the temporal coherence of the blurred regions and 
could result in distracting flickering of the blurred regions.

\textbf{Multi-level Security:} The coprime blur scheme described constructs two blurred video streams. One video stream is publicly accessible but access to the second stream is restricted to users with high clearance and is effectively private. Our scheme is therefore characterized by a public/private encryption/decryption pair.  Either of the blur kernel can act as a public encryption key whereas both blur encrypted blur streams act as a private key and as a result restricting access to both streams is paramount.  Our approach can be extended to implement a scheme that has more than two levels of security. We do this by dividing the users with access to the private video stream into levels. We restrict a level's access to the latent image by reducing the precision of the bits representing incoming blurred frames. For instance users with the highest access will have full bit resolution while others with lower clearance will not be able to access some subset of the least significant bits. Restricting access 
to the least significant bits of a pixel will therefore degrade the quality of the latent image reconstruction.

be more prudent to first blur the input with one coprime blur pair and use the resulting public video stream as the input to the second coprime blur pair. This second blur pair can be safely distributed in unsecured environments while knowing that the maximum level of recoverable detail is limited by the blur of the first coprime blur public video stream.

\section{Coprime Deblurring}
\label{coprime_deb}

\begin{figure*}
 \begin{center}
 \includegraphics[width=0.95\linewidth]{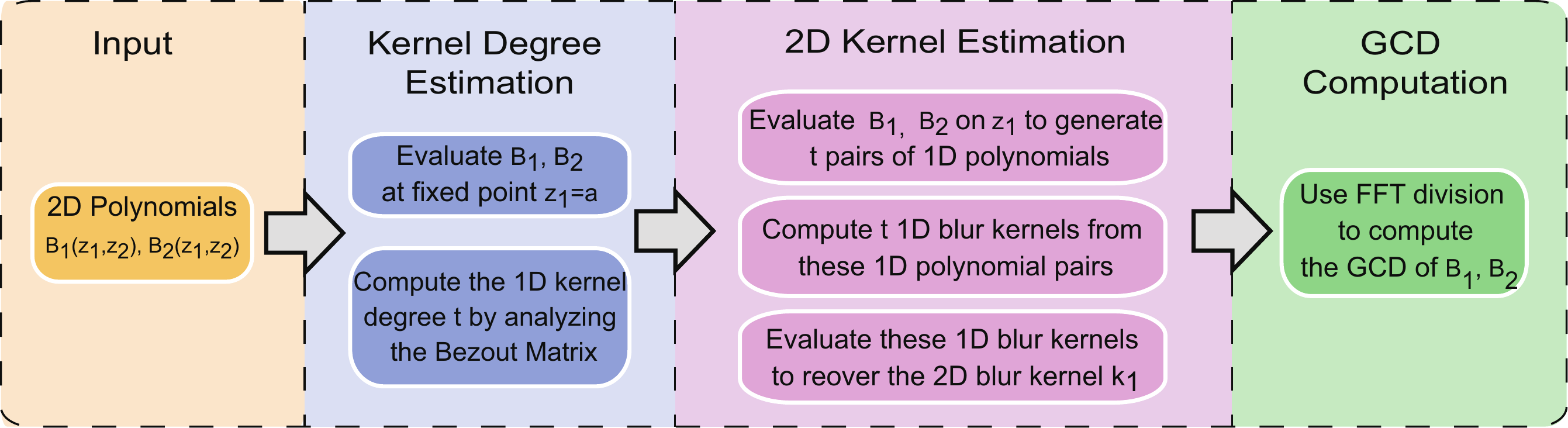}
 \end{center}
 \caption{GCD computation pipeline.}
 \label{fig:gcdpipe}
\end{figure*}

An image can be viewed as a bivariate polynomial with powers of x and y indicating horizontal and vertical positions and pixel values as coefficients. Some blurring operations then take the form of polynomial multiplication or convolution: we can model a blurred video frame $B$ of size $M \times N$ as the convolution of the input blur-free video frame $L$ hereafter referred  to as the latent image with an \emph{unknown} blur kernel $K$, plus noise $N_e$,
\begin{equation}
 B = L \otimes K + N_e,
\label{eq-blur-model}
\end{equation}
where $\otimes$ is the convolution operator. We then apply the z-transform to both sides of the equation. The z-transform of the image will have as coefficients the respective pixel values at each location. It allows the blurring process to be implemented as polynomial multiplication instead of convolution, 
\begin{equation}
 b(z_1,z_2) = l(z_1,z_2) \cdot k(z_1,z_2) + n_e(z_1,z_2),
\end{equation}
\noindent where $b$,$l$,$k$ and $n_e$ are z-transform of $B$,$L$,$K$ and $N_e$ respectively.
The two dimensional z-transform process can be further simplified by treating it as a pair of one dimensional transform,
\begin{equation}
 b(z_1,z_2) = \mathbf z_1^T \cdot B \cdot \mathbf z_2,
\end{equation}
\noindent where $\mathbf{z_1} = [1, z_1, z_1^2, \ldots, z_1^{M-1}]^T$ is the z-transform in one direction,
and $\mathbf{z_2} = [1, z_2, z_2^2, \ldots, z_2^{N-1}]^T$ is the z-transform in the other. 

Our proposed dual blur surveillance blurs the latent (all-clear) video frame using two \emph{coprime} blur kernels. When presented with two motion blurred images of the same scene $B_1$ and $B_2$ and neglecting the image sensor noise $N_e$, we could faithfully reconstruct the latent image $L$ together with the blur kernels $K_1$ and $K_2$ with respect to each blurred image as:
\begin{equation}
 l(z_1,z_2) = \textrm{GCD}\{b_1(z_1,z_2),b_2(z_1,z_2)\},
\end{equation}
where GCD corresponds to the greatest common divisor of $b_1$ and $b_2$. We also say that $k_1(z_1,z_2)$ and $k_2(z_1,z_2)$ are cofactors.

Fig.\ref{fig:gcdpipe} shows the details of our coprime deblurring algorithm. To recover the approximate GCD of $b_1$ and $b_2$, i.e. the latent image $L$, we first compute the approximate 2D cofactor $k_1$, and then directly divide the blur image $B_1$ by the kernel $K_1$ in the Fourier domain, since our estimated cofactor $k_1$ is numerically stable and accurate. The first step of our algorithm is to estimate the degree of the blur kernels. Here we assume the blur kernels are all square. We recover the kernel size by analyzing the singularity of the leading principal submatrices of the B\'{e}zout matrix \cite{Barnett:1972}.  We use this kernel degree $t$ in the second stage for computing the degree of the GCD and polynomial evaluation around the unit circle. Next we recover the 2D cofactor $k_1(z_1,z_2)$. Instead of solving a very large 2D GCD problem, we actually decompose it into $t$ 1D GCD sub-problems. It not only improves the computational efficiency dramatically, but also reduces the required memory 
space. Recall that the blur kernel size in our case is usually much smaller than the image resolution, therefore, our method is much more efficient than the conventional method \cite{Pillai99} based on recovering the latent image through GCDs. 

In practice, to solve different scaling factors with regard to each 1D GCD, we run the second step two times. We evaluate the 2D polynomials $b_1$ and $b_2$ along the $z_1$ direction first, compute the 1D GCDs, and then evaluate the 1D GCDs along $z_2$. Alternatively, we do polynomial evaluation along $z_2$ first and then $z_1$. The final estimation $k_1$ is the least square estimation of the two results generated by different polynomial evaluation orders. For technical details of our coprime deblurring algorithm, please refer to \cite{fli11,zijia10}.

\section{CUDA Implementation}
\label{CUDA_imp}

\begin{figure}
 \begin{center}
 \includegraphics[width=0.9\linewidth]{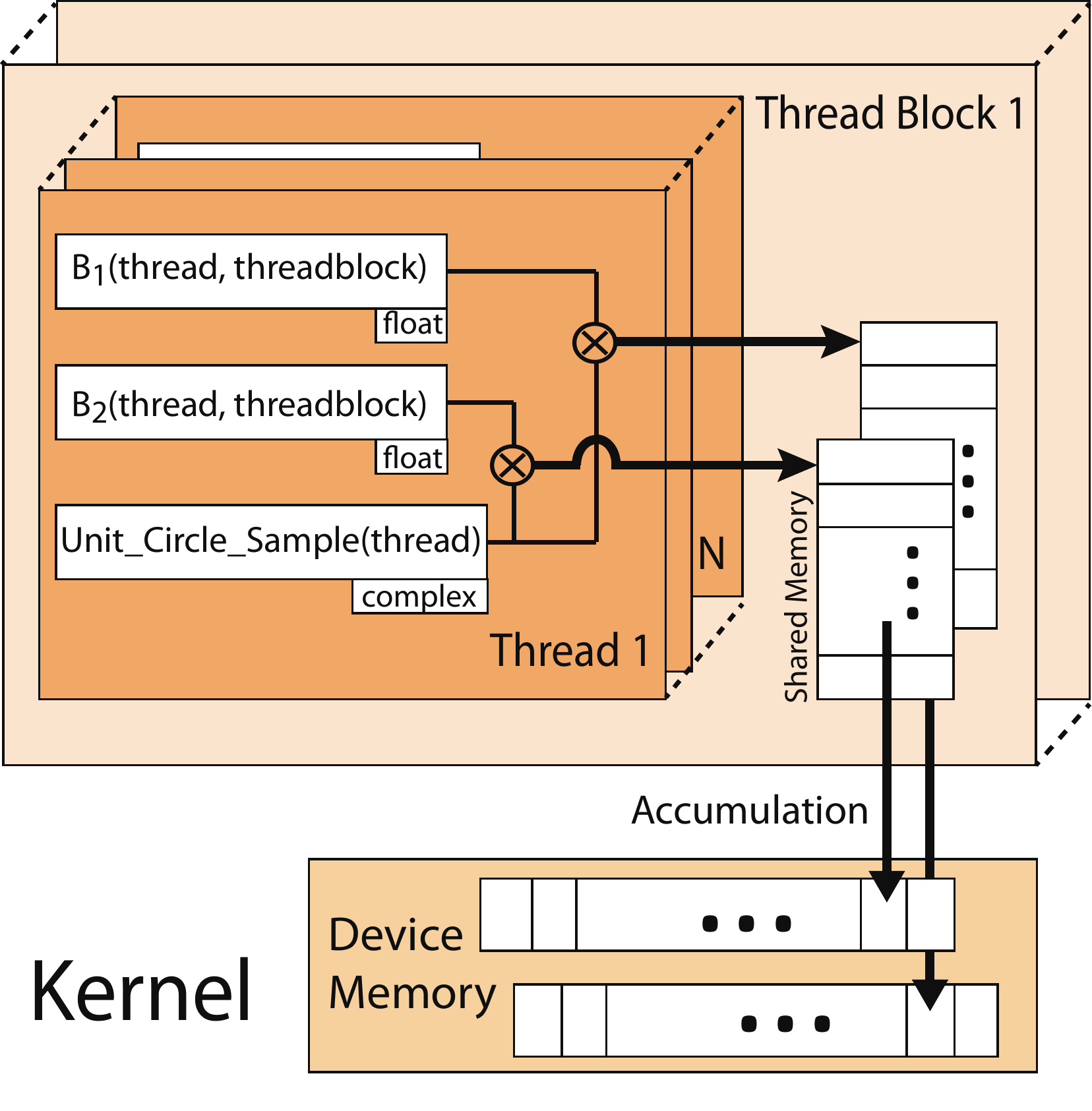}
 \end{center}
    \caption{CUDA device map for polynomial evaluation. Note that each thread only performs two multiplications and that results are written directly into shared memory. After accumulation of shared memory the result is written back to device memory.}
 \label{fig:cuda}
\end{figure}

The software implementation of our coprime deblurring is composed of five major stages, polynomial evaluation, kernel degree estimation, 1D cofactor estimation and 2D kernel estimation and inverse FFT. The processing pipeline is executed primarily on the GPU, however, some algebraic operations on small matrices such as SVD are implemented on the CPU. It was found that for small kernel sizes faster implementations for SVD exist on the CPU. Kernel degrees above 25 have a very detrimental impact on the running time and prevent the algorithm from running in real time. Therefore SVD is only performed on square matrices of degree no more than 25 in our implementation. We optimized the evaluation of the image polynomials around the unit circle and the final co-factor evaluations for efficient memory access. Portions of the processing pipeline which focused on producing smaller intermediate matrices were optimized primarily for maximizing throughput instead of minimizing the number of memory accesses.

The deblurring process begins with the kernel degree estimation process. In order to estimate the degree of the kernel, we first evaluate the two image polynomials at points around the unit circle in the $z_1$ direction or the $z_2$ direction. This process is exactly the N point Discrete Fourier Transform(DFT). We then choose the DFT coefficients corresponding to either the $z_1$ or the $z_2$ direction and construct a B\'{e}zout matrix. It does not matter which direction's coefficients are used first as we assume the blur kernel is square. We determine the degree of the blur kernel by finding the smallest leading principle submatrix that is non-singular. The general approach would be to use a binary search on the full B\'{e}zout matrix to find the smallest non singular submatrix. This approach however initially involves SVD on matrices too large to result in real time operation of our algorithm. Therefore, we restrict search to odd leading submatrices of size at least $9\times9$ and at most $25 \times 25$. 
As previously stated small SVD calculations are done on the CPU as they are better suited to take advantage of the cache on the CPU. In order to further speed the kernel degree estimation we use four threads to perform SVDs on different sized submatrices simultaneously. Of the 9 different odd kernels sizes between $9\times9$ and $25\times25$ it never takes more than two SVD calculations on each given thread to find the kernel degree. If all the B\'{e}zout submatrices are singular then the kernel degree must be maximal and 25 is chosen as the kernel degree $t$.

After determining the degree of the kernel we once again use the DFT coefficients but this time to compute 1D cofactor estimates. We now discuss our DFT implementation in greater detail before moving on to the 1D cofactor estimates. Evaluating the image polynomial in both the $z_1$ and $z_2$ direction required two calls to our CUDA DFT kernel.
Given a $t\times t$ kernel there will be $4t$ DFT calculations for the input blur pair. In our implementation, a $N$ point DFT in the $z_1$ direction on a $640\times480$ frame has 640 thread blocks each with 480 threads. Similarly, in the $z_2$ direction our kernel has 480 thread blocks and 640 threads per block. We have designed our DFT sampling kernel such that each thread only stores three values in the registers at any given time, as can be seen in Fig. \ref{fig:cuda}. Two of those values are the normalized intensity values from the same location in the CBP and the third value is the current complex sample from the unit circle. For a $t\times t$ kernel there will be $N$ samples of the unit circle in the $z_1$ direction DFT and $M$ different samples of the unit circle in the $z_2$ direction. Each thread performs two complex multiplications one for each image and stores the result in that thread block's shared memory array. The shared memory array is then accumulated to produce the DFT sample for that 
thread block after which the next unit circle sample is loaded from global memory into that register. The philosophy behind this approach is to ensure that any value used more than once resides either in a thread's register or shared memory within a thread block and is never loaded from global memory twice.

NVIDIA's CUDA provides support for the DFT operation via its cufft library. However, it does not provide good performance for input sizes that are not powers of two. For some frame sizes it can be impractical to pad input images to the next power of two considering the number operations that will be performed on the padded input. We therefore tailored our algorithm to provide the best DFT performance for any arbitrary frame size. Choosing a standard frame size does not restrict the DFT to that frame size as the level of privacy desired may increase or decrease the size of the blur kernel and therefore the resulting frame size. The number of points in our two 1D DFT sampling stage is therefore determined after kernel size estimation and a CUDA kernel function called to produce the correct unit circle samples.

Having performed the DFT on the image polynomials in the $z_1$ and $z_2$ direction we now begin the 1D cofactor computation by constructing $t$ B\'{e}zout matrices from $z_1$ and $t$ more B\'{e}zout matrices from the coefficients in the $z_2$ direction. Similar to the kernel degree estimation process these B\'{e}zout matrices are copied back to host memory to be factored by CPU SVD routines. The resulting 1D kernel cofactors are copied back to the GPU to be used in the 2D kernel estimation process. The GCD deblurring algorithm also has a number of intermediate processing steps on small matrices or vectors similar to the B\'{e}zout matrix. Unlike the previously discussed SVD decomposition these operations are all $O(n)$ and therefore present significant opportunity for performance gains. We do this by constructing composite matrices from vectors such as DFT coefficients and processing them in parallel across different thread blocks in order to maximize resource usage on the GPU.

2D kernel estimation continues by evaluating portions of the factored B\'{e}zout matrix on the unit circle to produce scaled coefficients of the kernel $k_1$'s Fourier transform. In order to combine the two estimates of $k_1$ we must recover the scaling factors on rows of the first $k_1$ estimate and the columns of the second $k_1$ estimate. We therefore have $2t$ unknowns and $t \times t$ elements in the kernel estimates producing an over determined system. We therefore employ a least squares approach and apply SVD again on the CPU to determine the scale factors. After normalization of the kernel estimates we compute the average of the two kernel estimates and use it to deblur one frame of input blur pair via the inverse FFT in CUDA's cufft library. We use CUDA's FFT implementation because it relies on texture memory and can store the entire image unlike our DFT evaluation kernel which is optimized for small number of DFTs.

\section{Results and Analysis}
\label{Res_ana}

\begin{figure*}
 \begin{center}
 \includegraphics[width=1\linewidth]{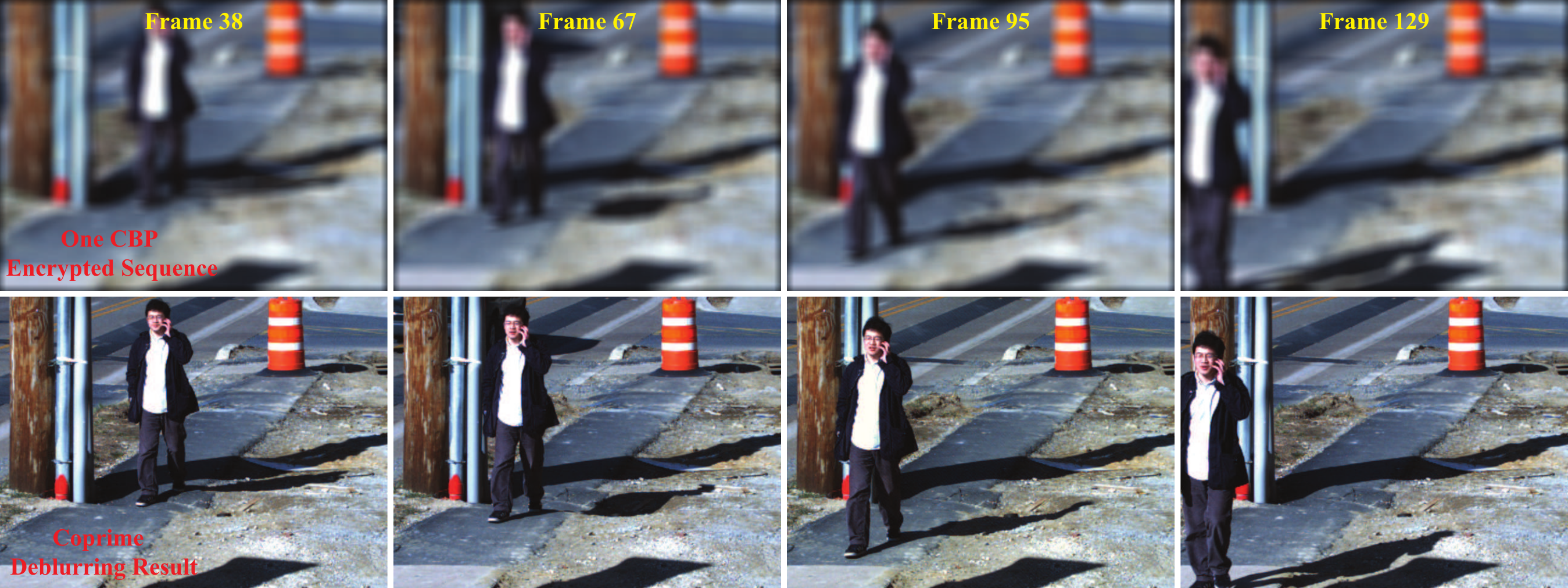}
 \end{center}
    \caption{Deblurring results of a typical surveillance scene in an urban environment.}
 \label{fig:feng}
\end{figure*}

\begin{figure*}
 \begin{center}
 \includegraphics[width=1\linewidth]{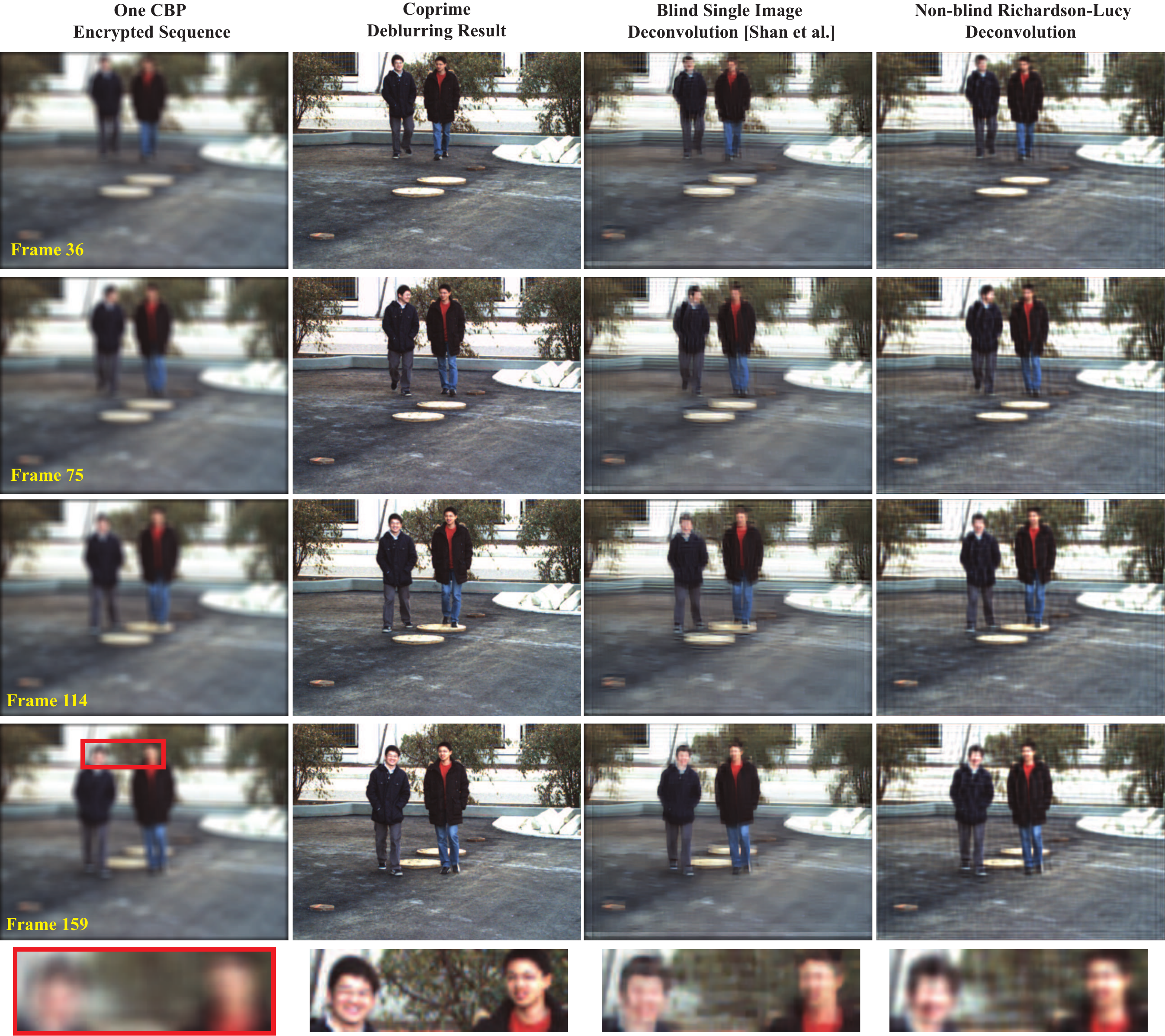}
 \end{center}
    \caption{Decrypted results with different deblurring methods on video frames encrypted by our coprime blurred scheme.}
 \label{fig:2pp}
\end{figure*}

In this section, we demonstrate our deblurring scheme by showing its performance in a number of common surveillance scenarios. We also present running times for our implementation for various blur kernel sizes. We test our algorithm on a PC with a 3.00 Ghz Intel Quad core CPU, 3 GB memory and a NVIDIA Geforce GTX460 GPU with compute capability 2.1. Video sequences were captured with a Point Grey Flea2 firewire camera. Our blurring and deblurring algorithms were implemented using NVIDA CUDA SDK 3.2. All input frames had a pixel resolution of $640\times480$. Blur kernel sizes varied between $19\times19$ and $25\times25$ based on the level of anonymity required for the scene. The effect of large kernels on the running time of our algorithm was also evaluated.

In Fig. \ref{fig:feng} we show a typical outdoor public surveillance scenario of a sidewalk in an urban area. Note that the frames blurred with $23 \times 23$ kernels in the top row  still allow the viewer to discern important qualities about the individual in the frame. Despite not being able to identify facial features one is still able to ascertain the color and type of clothing being worn. In addition, one can see that the individual has his hand raised to his ear and thereby infer that he might be making a telephone call without necessarily being able to see the mobile phone. In this way authorities tasked with surveillance are still able to identify common behavior without having being presented with a sharp image.

Note also in the bottom row samples of our deblurred results for this scene. In the event that greater scrutiny of the passerby is warranted a high clearance user would now be able to discern much more about the target. In addition to confirming that the passerby was indeed using a mobile phone they could determine that he was wearing glasses and that his  shoes had white tips. In urban surveillance scenes like this it is also common for cameras to inadvertently record the licence plates or passing cars. Notice the right lane of the street is within the camera's field of view. Employing our GCD blurring scheme in these sidewalk/street scenes would therefore restrict licence plate information to users with high clearance.

In Fig. \ref{fig:2pp} we present a parking lot surveillance video in which two distant individuals are walking towards the camera. The blur kernel in this series of frames is $19\times19$. We compare our deblurring results with those from the stat-of-the-art blind single image deblurring \cite{shan2008high}, and the traditional non-blind Richardson-Lucy deconvolution \cite{rl73}. Here non-blind deconvolution means we provide the ground truth blur kernels for the traditional Richardson-Lucy algorithm. As can be seen in the figure, both the blind single image deblurring algorithm \cite{shan2008high} and non-blind Richardson-Lucy deconvolution \cite{rl73} cannot recover the sensitive facial information from the subjects. However, in the second column our GCD deblurred frames have a much greater amount of detail and facial detail in particular. Only in the column presenting our GCD results are the details of the target's distant faces discernible.

\begin{figure*}
 \begin{center}
 \includegraphics[width=1\linewidth]{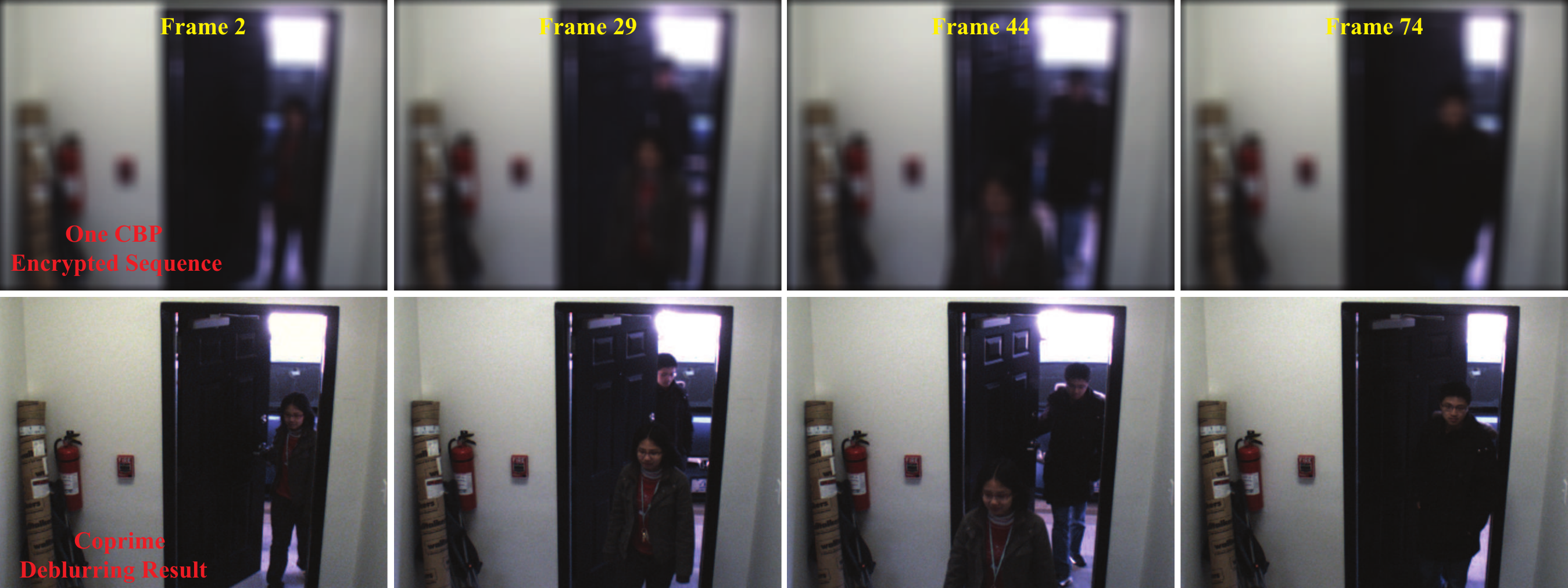} 
 \end{center}
    \caption{Deblurring results of a door entrance surveillance video sequence. } 
 \label{fig:jinwei}
\end{figure*}

In Fig. \ref{fig:jinwei} we show a camera monitoring an entrance a very common indoor surveillance scenario. Because objects of interest in the scene will be closer to the camera the size of the blur kernel was increased to $25 \times 25$ so as not to preserve too much detail. In the blurred frames it is clear that two people enter the building and what color their clothing is. However, apart from that it is difficult to discern facial details or even the gender of the individuals entering the building. In the deblurred frames on the bottom row it is easy to discern important details about the individuals such as their faces and even that there is a car in the background.

\begin{table}
\centering
\caption{Running times of the pipeline stages with different kernel size for images of size $640 \times 480$ (in milliseconds) }
\label{tab:runningtime}
\begin{tabular}{c|c|c|c} \hline
\hline
\multirow{2}{*}{Pipeline Stage} & \multicolumn{3}{c}{Kernel Size} \\ \cline{2-4}
 & $9 \times 9$ & $17 \times 17$ & $23 \times 23$  \\ \hline
Polynomial Evaluation & 0.65 & 0.94 & 1.39\\ 
Kernel Degree Estimation & 0.81  & 0.81 & 0.83\\ 
1D Kernel Estimation & 18.45   & 27.83 & 35.87\\ 
2D Kernel Est. and FFT& 4.06  & 4.82 & 5.9\\ \hline 
Total Time & 23.97 & 34.4 & 43.99 \\
\hline\hline
\end{tabular}
\end{table}

\textbf{Performance}. Tab. \ref{tab:runningtime} provides running times for our overall algorithm and component processes with various kernel sizes. From the total running times it is evident that our processing pipeline achieves frame rates that approach or exceed current standard video frame rates. Running times include the memory copies of the input frames to the GPU and the memory copy of the resulting unblurred frame back to host memory. We measure the running times using calls to NVIDA's timer functions. Table 1 shows that of all the processing pipeline stages the 1D kernel estimation is the most time consuming. This is primarily due that time associated with each SVD decomposition and the fact the number of SVD factorizations grow exponentially with kernel size. However, other factors related to the construction of a greater number of Bezout matrices, copying them to host memory and copying the result of the SVD back to GPU scale in proportion to the number of SVD factorizations.

Less computationally expensive stages such as polynomial evaluation and kernel degree estimation contribute little to the overall running time of the algorithm. However, the polynomial evaluation running time increases proportionally with kernel degree $t$. The running time of the kernel degree estimation stage remains essentially constant as the size of the input remains the same regardless of the blur kernel size. The running time of the last stage however, increases steadily with kernel size. This increase is due to the small polynomial evaluation carried out on the 1D kernel cofactors to produce the 2D kernel estimate and not to the constant running time to divide by the kernel and the apply the the inverse FFT to deblur the image. We also execute our algorithm on the CPU producing running times of $\sim$0.509 secs and $\sim$3.404 secs for $9\times9$ and $25\times25$ blur kernels respectively. Running the vast majority of our algorithm on the GPU therefore produces speedups of 21 and 77 respectively.

\section{Conclusions and Future Work}
\label{conc}


We have demonstrated the usefulness of our coprime blur scheme for visual data hiding in surveillance that does not rely on cryptography to completely obscure the images contents. Instead the ability to recover all of the image information relied on access to a second stream of images not a predetermined encryption key. Only with access to the second stream is one able to recover the blur kernel and finally the latent image. This key-less decryption allows a multi-tiered security clearance system to be implemented by structuring or modifying the video transmission infrastructure. We have implemented a GPU based processing pipeline that produces deblurred video at frame rates above 20fps for relatively large kernel sizes upto $25\times25$.

Our GPU implementation of the GCD image deblurring algorithm has several limitations. First, our approach requires double the bandwidth for video transmission and a means to restrict access to both video streams. These requirements may reduce the number of instances where deploying our approach is practical.
The running time for our approach is proportional to the blur kernel size and therefore the level of privacy in some surveillance scenarios may be limited by a desired frame rate. In addition the running time of the deblurring process may prohibit some post processing that may aid surveillance such as facial or object recognition. Future work will include decreasing the running time of expensive operations in the pipeline to facilitate some post processing steps such as background subtraction. Reducing the running time can be achieved either by optimizing the existing pipeline or introducing a hybrid pipeline. For example shifting computation of SVD to an external FGPA or coprocessor. Another topic of interest we may pursue is the integration of a z-camera which will provide the algorithm with depth information which then can be used to compute the optimal blur kernel size for hiding information.


\bibliographystyle{spmpsci}      

\bibliography{egbib}  

\end{document}